# SSL-NBV: A Self-Supervised-Learning-Based Next-Best-View algorithm for Efficient 3D Plant Reconstruction by a Robot

Jianchao Ci[a,*], Eldert J. van Henten[a], Xin Wang[a], Akshay K. Burusa[a], Gert Kootstra[a]

[a]*Agricultural Biosystems Engineering Group, Department of Plant Sciences,*
*Wageningen University and Research, P.O. Box 16, Wageningen, 6700AA,*
*the Netherlands*

## Abstract

The 3D reconstruction of plants is challenging due to their complex shape causing many occlusions. Next-Best-View (NBV) methods address this by iteratively selecting new viewpoints to maximize information gain (IG). Deep-learning-based NBV (DL-NBV) methods demonstrate higher computational efficiency over classic voxel-based NBV approaches but current methods require extensive training using ground-truth plant models, making them impractical for real-world plants. These methods, moreover, rely on offline training with pre-collected data, limiting adaptability in changing agricultural environments. This paper proposes a self-supervised learning-based NBV method (SSL-NBV) that uses a deep neural network to predict the IG for candidate viewpoints. The method allows the robot to gather its own training data during task execution by comparing new 3D sensor data to the earlier gathered data and by employing weakly-supervised learning and experience replay for efficient online learning.

Comprehensive evaluations were conducted in simulation and real-world environments using cross-validation. The results showed that SSL-NBV required fewer views for plant reconstruction than non-NBV methods and was over 800 times faster than a voxel-based method. SSL-NBV reduced training annotations by over 90% compared to a baseline DL-NBV. Furthermore, SSL-NBV could adapt to novel scenarios through online fine-tuning. Also using real plants, the results showed that the proposed method can learn to effectively plan new viewpoints for 3D plant reconstruction. Most importantly, SSL-NBV automated the entire network training and uses continuous online learning, allowing it to operate in changing agricultural environments.

## 1. Introduction

Greenhouse production is a widely used method for vegetable cultivation, offering substantial benefits such as an extended plant production period, improved quality, and quantity, and finally leading to increased profitability for growers. The selection of suitable cultivars, i.e. phenotyping, as well as the crop production process itself, including tasks like harvesting and de-leafing, require a significant amount of labor. This limits large-scale greenhouse production because experienced labor is scarce and labor costs are high. Automating these labor-intensive tasks using robotics can help reduce reliance on human labor and potentially increase greenhouse productivity and production efficiency.

Robotic phenotyping and greenhouse operations like harvesting and leaf removal, are essentially based on the same key functionality, i.e. perception. However, accurate robotic perception in complex greenhouse environments is challenging due to the presence of occlusion and variation. Occlusion is mainly caused by leaves, preventing sufficient information collection, thus increasing uncertainty in robotic operations. Variations arise from dynamic growing environment (e.g., lighting) and inherent plant properties (e.g., shape, size, and texture), making the systems may work in a certain condition but lack robustness to variation. The objective of this study was to develop a robust robotic perception system to capture sufficient information of the plants under the presence of occlusion. The system was specifically developed and evaluated in plant phenotyping scenario for plant reconstruction tasks, but with the potential to be used for other tasks such as robotic harvesting after small modifications and fine-tuning.

* Corresponding author.
E-mail address: jianchao.ci@wur.nl.

Plant phenotyping is a set of methodologies to measure the plant traits, such as leaf angle, node length, and leaf area, and subsequently link these measurements to plant genotypes and growing environments, which is a key technology in plant breeding for efficient selection of target cultivars (Poland and Rife, 2012). Compared to traditional manual assessment of plant traits, robotic phenotyping methods have received significant interests by integrating machine vision techniques and robotics, showing potential of plant phenotyping in an automatic, accurate, non-destructive, and high-throughput manner (Atefi et al., 2019; Hartmann et al., 2011; Polder and Hofstee, 2014; van der Heijden et al., 2012). Two-dimensional (2D) robotic phenotyping methods have been widely studied (Jansen et al., 2009; Minervini et al., 2014; Tisné et al., 2013). However, a significant disadvantage of these methods is they cannot accurately measure plant traits that are expressed in three dimensions (3D) such as plant volume. Alternatively, three-dimensional (3D) methods perform measurement in 3D data, offering more comprehensive and accurate information of plant traits (Boogaard et al., 2023; Shi et al., 2019). An essential first step of 3D phenotyping methods is generating a 3D digital reconstruction of the plants, which can be represented as a point cloud (Boogaard et al., 2022, 2021), or a mesh (Thapa et al., 2018; Vázquez-Arellano et al., 2018). However, accurate 3D plant reconstruction is extremely challenging due to significant plant occlusion. A common approach to solve this problem is using multiple viewpoints (Golbach et al., 2016; Lu et al., 2017; Shi et al., 2019), whereby the information that is occluded in a single view becomes available. However, as these methods employ a passive paradigm, where data collection is constrained to predefined views and trajectories, they suffer from missing data, redundant information collection, and the need for continuous fine-tuning for novel plants.

Recently, Next-Best-View (NBV) methods in agricultural robotics have gained significant interest (Gibbs et al., 2020, 2018; Zaenker et al., 2021a; Zapotezny-Anderson and Lehnert, 2019). NBV adopts an active perception paradigm to actively reposition and reorient the camera. The selection of the next viewpoint relies on analyzing the observed data so far, considering the objective at hand, with the aim of maximizing the information gain (IG). IG is a metric that quantifies the increase in information achievable by choosing a new viewpoint. This approach enables more efficient and flexible data collection compared to passive perception methods (Atefi et al., 2021). Accurate and efficient estimation of IG is crucial for NBV methods. Traditional NBV methods require maintaining an occupancy grid representing space in voxels and corresponding occupancy probabilities indicating whether a voxel is free or occupied, we call this Voxel-NBV. IG estimation of a candidate viewpoint is performed by casting rays from the viewpoint into the occupancy grid and counting the voxels (both free and occupied) that the rays traverse. This method has been employed in various agricultural applications (Burusa et al., 2022; Gibbs et al., 2020, 2018; Zaenker et al., 2021b). This method is useful as it estimates IG explicitly, making the planning process interpretable, and provides information about both occupied and unoccupied space. However, it can be computationally and memory-intensive due to casting millions of rays and examining the transitions of each ray at a voxel level.

Deep-Learning-based Next-Best-View (DL-NBV) methods aim to address these downsides. These methods use neural networks to directly predict the IG based on the observed data so far, eliminating the need for computationally expensive ray-casting and voxel-wise operations. For example, Mendoza et al. (2020) proposed NBV-Net, which utilizes a 3D convolutional neural network to directly predict the IG of a set of pre-defined candidates using the accumulated occupancy grid as input. Similarly, Zeng et al. (2020) proposed a point-cloud-based neural network called PC-NBV, which predicts the

* Corresponding author.
E-mail address: jianchao.ci@wur.nl.

IG values of all the candidates by taking the accumulated point cloud as input. PC-NBV offers higher planning efficiency by eliminating the step of converting the point cloud to a volumetric map. While both methods display high reconstruction efficiency and robustness to man-made objects with simple geometries, their performance in reconstruction complex plant structure has not been evaluated. Moreover, both methods require separate training steps and complete object models for ground-truth IG estimation, posing challenges for use in dynamic agricultural environments. In this paper, we will develop a DL-NBV method based on the existing PC-NBV method but with the capability of online self-supervised learning and evaluate its performance in an agricultural scenario focusing on plant phenotyping.

Training DL-NBV networks for plant reconstruction poses a significant challenge due to the difficulty of acquiring large amounts of IG-labeled training data. Manual calculation of ground-truth IG is impractical as it is subjective and difficult to accurately measure IG values. However, using a robot, this process can be automated by having the robot explore the environment. Thus, the robot can autonomously collect and annotate IG for data for training even during execution of the task, enabling to improve performance in a life-long and self-supervised manner. This learning approach is termed Self-Supervised Learning (SSL) (Deng et al., 2020). A challenge of SSL is the automatic annotation of IG. Mendoza et al. (2020), Wang et al. (2019), and Zeng et al. (2020) compute ground-truth IG of a candidate view by comparing to the complete model. However, these methods require prior knowledge of the object shape, making them infeasible for agricultural applications as plant models are not available in advance. To solve this problem, this research proposes an improved IG calculation metric based on the method of Zeng et al. (2020) but enables ground-truth IG calculation solely based on the data collected by the robot, thereby achieving self-supervised learning.

Another problem that prevents the application of current DL-NBV methods in agriculture is their reliance on offline training, i.e., initially collecting large amounts of training data, training the method and then execution in a separate step. This limits the method to continue to improve or to adapt to novel targets and environments (Kahn et al., 2020). Conversely, online learning algorithms integrate both processes into a continuous feedback loop, allowing the network to learn and optimize while executing the task, facilitating adaptation to new targets and environments. However, online learning methods often struggle with low sample-efficiency, as they focus on the most recent data and discard historical data (Zhang and Sutton, 2017). To address the challenges of online learning, Mnih et al. (2013a) employed an off-policy training method called 'experience replay' in the field of Reinforcement Learning (RL), demonstrating improved sample-efficiency and generalization (refer to step 5 of section 2.1.1 for detailed explanation). This inspired our work. Additionally, existing DL-NBV methods typically undergo training with strongly-supervised data, wherein each training input requires IG-annotation for all candidates as ground-truth, leading to inefficient data collection. To address this issue, we employed a weakly-supervised learning technique to reduce the number of required IG annotations, facilitating efficient training data collection and online learning. Further discussion is provided in section 2.1.4.

The objective of this paper is to propose and evaluate a novel SSL-NBV method for 3D robotic plant reconstruction tasks. Specifically, the state-of-the-art learning-based NBV method is improved to enable online SSL of IG prediction, allowing for online network adaptation to novel plants during task execution. To achieve successful online self-supervised learning, an IG metric is designed to eliminate the need for complete plant models in ground-truth IG calculation, an experience replay technique is employed to improve sample-efficiency, and weakly-supervised learning is utilized for efficient training data collection. The method was tested in both simulation and real-world scenarios, providing a comprehensive evaluation in comparison with various baseline planners. The real-world experiments demonstrated that the proposed method is also capable to learn view planning for the reconstruction of real plants. The study addressed the following questions:

1. What reconstruction quality and efficiency can the proposed SSL-NBV achieve in comparison with other NBV (PC-NBV and Voxel-NBV) and non-NBV (Random and Pre-defined) methods?
2. How does the proposed SSL-NBV, using online and weakly-supervised learning, compare to PC-NBV, using offline and strongly-supervised learning?
3. Can SSL-NBV adapt to a new environment during online self-supervised learning?
4. Can the proposed SSL-NBV approach be applied in a noisy real-world scenario?

* Corresponding author.
E-mail address: jianchao.ci@wur.nl.

## 2. Methods and materials

This section describes the method and experimental setup. Section 2.1 describes the online self-supervised learning process for plant reconstruction. The experimental setup, including simulation and real-world scenarios, is detailed in section 2.2. Finally, section 2.3 presents the metrics used for evaluating the performance of the proposed method.

### 2.1. Online self-supervised learning for plant reconstruction

The entire learning process involves continuous iterations. At each iteration, the robot performs next-best-view planning using a deep neural network to select a new camera pose from a set of candidate viewpoints to collect new data for 3D plant reconstruction, which simultaneously generates new training data to update the network. This online learning process allows the network to be updated during task execution and without the need for a separate training phase. This allows the robot to continuously improve and to adapt to a changing environment. The algorithm was first developed and validated in the simulation for repid development and repeated experiment, then evaluated in the real-world application.

The entire learning process consists of a maximum number of $T$ iterations. The candidates set $C = \{c_1, ..., c_M\}$ is defined before the entire learning process, where $c_i \in \mathbb{R}^6$ is a camera pose consisting of position and orientation $\{x_i, y_i, z_i, \alpha_i, \beta_i, \gamma_i\}$, and $M$ is the total number of candidate views. Briefly, the learning process works by iteratively selecting views from $C$ as camera pose for plant reconstruction and collecting data to update the network. View change and data collection are conducted using a robot arm with a depth camera mounted as end effector. From each view, a partial point cloud is collected, and the point clouds collected from all views are fused as a digital plant reconstruction.

Fig. 1 provides an overview of an iteration. An iteration $t$ consists of seven steps: (1) Using a DL-NBV network to predict the IG, $\hat{G}_t$ by taking as input the current accumulated point cloud $P_t^a$ and a view-selection state $V_t$ keeping track of previously visited viewpoints; (2) selecting the next view $v_{t+1}$ based on $\hat{G}_t$; (3) moving the camera to $v_{t+1}$ and collecting a new partial point cloud $P_{t+1}^c$; (4) calculating the actual IG $G_t$ based on $P_{t+1}^c$ and $P_t^a$; (5) storing supervision pair with $P_t^a$ and $V_t$ as the input, and $G_t$ as the ground-truth; (6) updating the accumulated point cloud $P_{t+1}^a$ and view-selection state $V_{t+1}$; and (7) training the network weights $\mathbb{W}_{t+1}$ of the DL-NBV based on a batch from the training data. The steps are explained in detail in the next subsection.

* Corresponding author.
E-mail address: jianchao.ci@wur.nl.

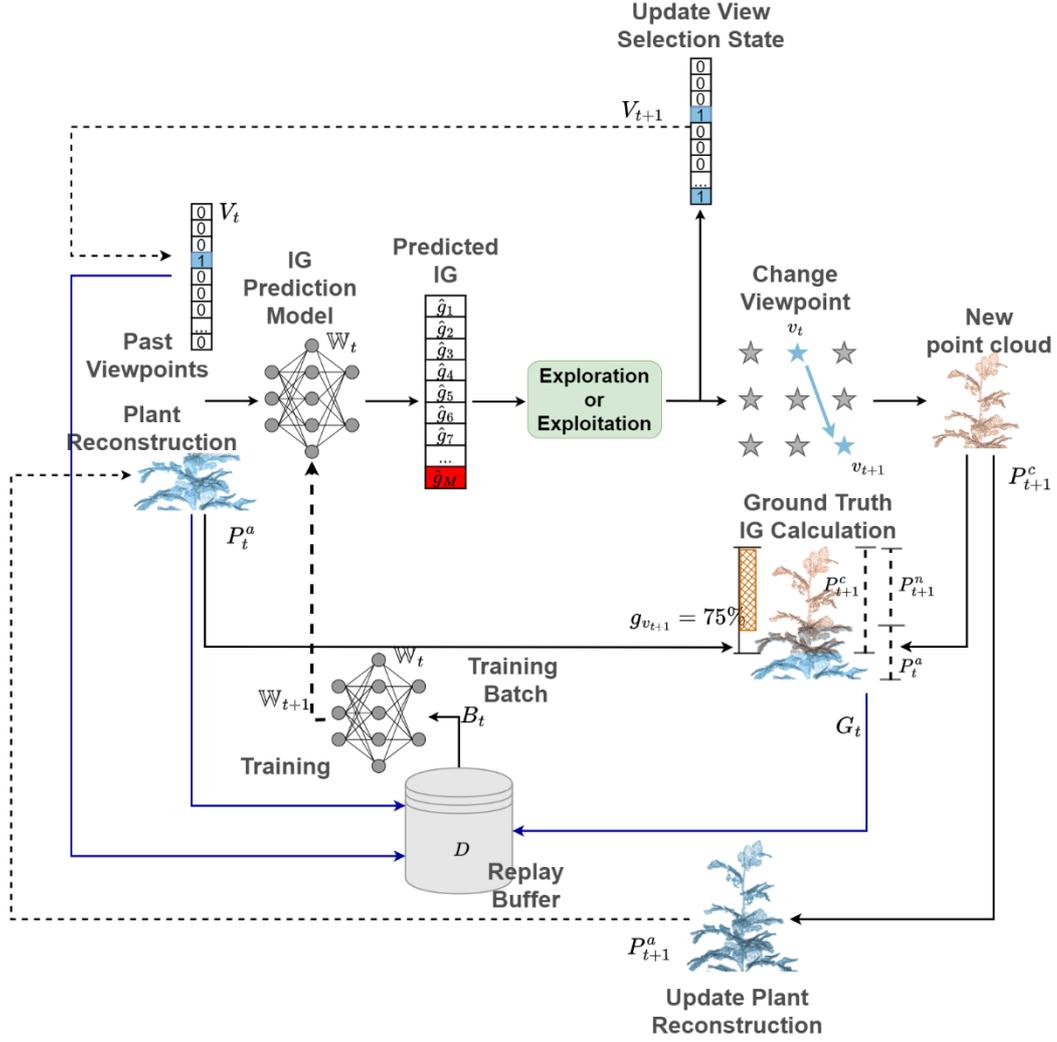

Fig. 1. Illustration of an iteration step in self-supervised learning. At each iteration step $t$, next-best-view planning is conducted for plant reconstruction, and generated data, including the accumulated point cloud, view selection states, and ground-truth IG, are collected to optimize the network's weights. Blue lines represent the collection of training samples, dotted black lines represent data updates, and solid black lines represent data transformation after the operation.

### 2.1.1. The iteration steps in detail

In step 1, the current accumulated point cloud $P_t^a$ and view-selection state $V_t$ are fed into the DL-NBV network to predict $\hat{G}_t = \{\hat{g}_1, \dots, \hat{g}_M\}$, which contains IG for all candidates, where $\hat{g}_i \in \mathbb{R}$ represents the predicted IG for a viewpoint. $P_t^a = \{p_1, \dots, p_K\}$ is the accumulated point cloud containing all collected points, where $p_i = \{x_i, y_i, z_i\} \in \mathbb{R}^3$ is a 3D point with xyz-coordinates. $V_t = \{s_1, \dots, s_M\}$ is a vector containing the selection state for all candidate views, with $s_i \in \{0,1\}$ equaling 1 if a viewpoint was selected in the past and 0 otherwise. At the beginning of the learning process, $P^a = \emptyset$ and $V = \{0, \dots, 0\}$. In this research, we used the PC-NBV network (Zeng et al., 2020) as the DL-NBV planner. A detailed explanation of the PC-NBV network structure and IG prediction is provided in section 2.1.2.

In step 2, based on $\hat{G}_t$, the estimated next-best view of iteration $t$ is determined as $\hat{v}_{t+1} = \text{argmax}_{j=1}^{M} \hat{g}_j$, corresponding to the view acquiring the highest predicted IG. However, instead of using $\hat{v}_{t+1}$ directly as the next camera view $v_{t+1}$, an exploration mechanism is employed to maintain an exploration-exploitation balance during online learning. Exploitation involves selecting $\hat{v}_{t+1}$ as the next camera view, which prevents wasteful exploration, while exploration entails randomly picking a viewpoint $v^r$ from the candidate set to avoid the network getting stuck in a suboptimal view selection policy. We devised an exploration mechanism following a commonly used paradigm as in Wang et al. (2016). The mechanism is governed by the exploration rate $\epsilon_t \in [\epsilon_{min}, \epsilon_{ini}]$. At the beginning of the entire training process, when $t=1$, $\epsilon_t = \epsilon_{ini}$ was set as the initial exploration rate. Then, $\epsilon_t$ is decayed by a decay rate $\rho$ every iteration until reaching a minimum $\epsilon_{min}$, according to $\epsilon_t = \max(\epsilon_{min}, \rho^{t-1} \cdot \epsilon_{ini})$. In this work, we used $\rho=0.95$ as a constant. The next viewpoint is then determined as follows:

$$v_{t+1} = \begin{cases} \hat{v}_{t+1}, & \text{if } \epsilon_t < x \sim U(0,1) \\ v^r \in_R C, & \text{otherwise.} \end{cases} \quad (1)$$

\* Corresponding author.
E-mail address: jianchao.ci@wur.nl.

If a randomly generated number $x$ from a uniform distribution with bounds 0 and 1 is greater than $\epsilon_t$, the next viewpoint $v_{t+1}$ is set to $\hat{v}_{t+1}$, otherwise, it is randomly selected from the candidate set $C$. This exploration mechanism enables the system to perform extensive exploration in the beginning and gradually shifts towards using self-collected information as the network improves from experiences.

In step 3, after the next view is determined, the camera pose is changed to this view, allowing for the collection of a new partial point cloud $P_{t+1}^c$. A detailed explanation of point cloud collection and processing is provided in section 2.1.3.

In step 4, based on $P_t^a$ and $P_{t+1}^c$, the ground-truth IG $g_{v_{t+1}}$ for viewpoint $v_{t+1}$ is calculated and is then converted into $G_t = \{0, \dots, g_{v_{t+1}}, \dots, 0\}$ as the target vector for network training. To achieve continuous online learning in real agricultural scenarios, the robot should be able to calculate $g_{v_{t+1}}$ from its own collected data. To this end, we propose a method enabling ground-truth IG calculation of a viewpoint based solely on $P_{t+1}^c$ collected from this view by robot and the previously collected data $P_t^a$. Briefly, $g_{v_{t+1}}$ is calculated as the proportion of points in $P_{t+1}^c$ that provide new information that was not captured in $P_t^a$. As shown in Fig. 2, the subset of newly observed points, denoted as $P_{t+1}^n$ (marked orange), within $P_{t+1}^c$ is obtained by removing the intersection (marked green) with $P_t^a$, calculated as $P_{t+1}^n = P_{t+1}^c - (P_{t+1}^c \cap P_t^a)$. Then, the ground-truth IG for view $v_{t+1}$ is calculated as $g_{v_{t+1}} = |P_{t+1}^n|/|P_{t+1}^c|$, where $|\cdot|$ denotes the size of the set of points.

However, intersection $P_{t+1}^c \cap P_t^a$ cannot strictly be calculated using the intersection of the sets, because the corresponding points in $P_{t+1}^c$ and $P_t^a$ will not have the exact same coordinates. Instead, we determine $P_{t+1}^c \cap P_t^a$ by calculating the Euclidean distance between each point in $P_t^a$ for each point in $P_{t+1}^c$. If the smallest distance is below a threshold $\delta$, the point is added to the intersection. So, the intersection is defined as:

$$P_{t+1}^c \cap P_t^a = \left\{ p_k \middle| p_k \in P_{t+1}^c \land \min_{p_j \in P_t^a} \|p_k - p_j\|_2 \leq \delta \right\} \quad (2)$$

where $\|\cdot\|_2$ is the Euclidean distance between two points. $\delta$ was set at 0.003m in the simulation, while set at 0.01m in the real word. This aligned with the voxel size used for point cloud downsampling (refer to section 2.1.3).

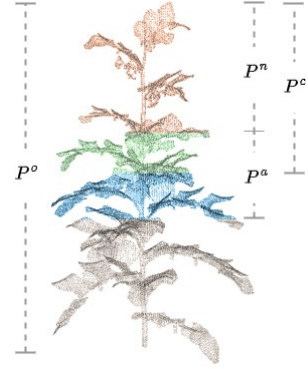

Fig. 2. Illustration of different types of point clouds. The ground-truth IG of a viewpoint equals the proportion of novel points with respect to the observed data from this view. $P^a$ (blue and green points) refers to the accumulated point cloud. $P^c$ (orange and green points) refers to the newly collected partial point cloud. $P^n$ (orange points) refers to the newly collected points within $P^c$. Green points refer to the intersection between $P^a$ and $P^c$. $P^o$ (grey points) refers to the complete point cloud, which is used in the evaluation phase.

This method allows for the calculation of ground-truth IG solely based on the data collected by the robot, facilitating an autonomous training data collection when integrated with the flexibility of a robotic system.

* Corresponding author.
E-mail address: jianchao.ci@wur.nl.

The formulated target vector $G_t$ for network training containing only the sparse ground-truth IG for a single view is called *weakly-supervised annotation*. The detailed explanation of using weakly-supervised data in network training is provided in section 2.1.4.

In step 5, the accumulated point cloud $P_t^a$ and the view-selection state $V_t$ are collected as inputs, along with the target vector $G_t$, to formulate a training sample $b_t = \{P_t^a, V_t, G_t\}$, which is then stored in a circular buffer $D = \{b_{t-l+1}, \dots, b_{t-1}, b_t\}$, where $l \in \mathbb{N}$ is the maximum number of samples that can be stored. When the capacity of $D$ is reached, the oldest data is replaced by the newest data, maintaining a fixed-size representation of past experiences.

In step 6, the accumulated point cloud and the view-selection state are updated. The accumulated point cloud is updated by adding the newly collected point cloud, $P_{t+1}^a = P_t^a \cup P_{t+1}^c = P_{t+1}^c - (P_{t+1}^c \cap P_t^a) + P_t^a$ (the calculation of $P_{t+1}^c \cap P_t^a$ refers to Eq.(2)), and the view-selection state is updated to $V_{t+1}$ by setting the $s_{v_{t+1}}$ to 1.

Finally, in step 7, a subset $B_t \subseteq_R D$ is randomly sampled to optimize the weights $\mathbb{W}_t$ of the network, where the subscript $R$ indicates a random selection of training samples. $B_t = \{s_1, \dots, s_N\}$ contains $N$ training samples, where $N \in \mathbb{N}$ is the batch size, and network training starts when the stored number of samples in $D$ exceeds the batch size. This method, collecting historical data in a buffer and then sampling data from the buffer to train the network, is called experience replay. Experience Replay is a key technique utilized in online learning to improve sample efficiency through the reuse of historical data and improve training stability by breaking temporal correlations between consecutive steps.

### 2.1.2. Deep-learning network structure for information gain prediction

Fig. 3 shows the architecture of the network used in this research, which follows the original paper in which PC-NBV was presented (Zeng et al., 2020). At any iteration step $t$, the network takes $P_t^a$ and $V_t$ as inputs. First, the point cloud $P_t^a$ is processed by a feature extraction network, extracting local features to generate the point-wise feature *F0* with 264 features per point. *F0* is further processed using max pooling to obtain the global feature *G0* of length 264. Subsequently, both *G0* and $V_t$ are duplicated to match the dimension of *F0* along the vertical axis. This duplication facilitates the concatenation with *F0*, resulting in the generation of point-wise features *F1*, containing both global and local features of the point cloud, as well as the view-section state. *F1* is then input to a self-attention unit (Zhang et al., 2019) to further integrate these features, yielding the attention feature *F2*. Following this, a multi-layer perceptron (MLP1) module with 2 layers of 1024 neurons each following max pooling are applied, producing *G1* as the final global feature. *G1* represents $P_t^a$ and $V_t$, which is finally fed into another MLP2 with 4 layers of 1024, 512, 256, and 33 neurons each. MLP2 predicts $\hat{G}_t = \{\hat{g}_1, \dots, \hat{g}_M\}$, containing the predicted IG for all candidate viewpoints.

Notably, the original PC-NBV is designed for offline learning and tested in a simulation environment. We modified the training approach, IG calculation metric (step 4 of section 2.1.1), and loss function (section 2.1.4), enabling continuous online learning with weakly-supervised data, showing its capability in real agricultural scenarios.

### 2.1.3. Point cloud collection and processing

In the simulation environment, a simulated Intel Realsense L515 RGB-D camera was utilized to capture both color and depth information from the viewpoint, which were then combined to form a partial point cloud. Point cloud downsampling was subsequently performed using the

* Corresponding author.
E-mail address: jianchao.ci@wur.nl.

VoxelGrid[1] filter with a voxel size of 0.003m, generating $P^c$. The choice of voxel size aligned with the work of Burusa et al. (2022), who employed a Voxel-based NBV method for plant reconstruction, enabling direct comparison with their approach. In the real-world experiment, an Intel Realsense L515 RGB-D camera was used to capture point clouds. Real-world point clouds commonly contain various sources of noise due to ambient lighting fluctuations, sensor-specific artifacts, and idiosyncrasies in point cloud generation algorithms. To address this, a three-step noise reduction process was implemented. Initially, a RangeFilter was employed to eliminate points falling outside a specified range. The cropped point cloud then underwent further refinement through the application of the Statistical Outlier Removal (SOR) filter, which identified and removed points significantly deviating from their neighbors when compared to the point cloud's average. Finally, the VoxelGrid filter with a voxel size of 0.01m was applied to further enhance the cleanliness of the point cloud data.

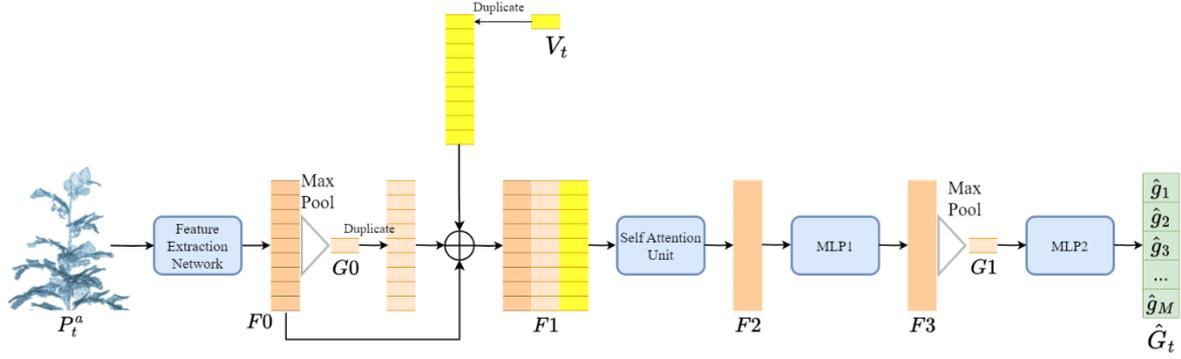

Fig. 3. The architecture of PC-NBV network. The network initially extracts the global feature of the accumulated point cloud and then combines it with the local feature and view selection state to predict the IG for each candidate viewpoint. The figure is adopted from Zeng et al. (2020).

### 2.1.4. Loss calculation using weakly-supervised data

In the original PC-NBV, a classic mean square error (MSE) loss function was utilized to calculate the loss $L^s$ (superscript $s$ represents the loss for strongly-supervised learning) between $\hat{G}_t$ and $G_t$. The $G_t$ was formulated as $G_t = \{g_1, \ldots, g_M\}$, encompassing the ground-truth IG for all candidate viewpoints. The $L^s$ was computed as follows:

$$L^s = \sum_{i=1}^{M}(g_i - \hat{g}_i)^2 \qquad (3)$$

This approach requires each $\hat{g}_i$ to be associated with a corresponding $g_i$ for supervision, or in other words, it demands fully ground-truth labels, using strongly-supervised learning. While valid in simulation environments, this method is inefficient for robotic online learning in real-world scenarios, as robotic motion is time-consuming and obtaining $G_t$ necessitates the robot traversing all candidate viewpoints in $C$.

To address this issue, we propose a weakly-supervised learning method, enabling training based on coarse-grained ground-truth labels, a specific technique called inexact supervision (Zhou, 2018). We defined ground-truth vector $G_t = \{0, \ldots, g_{v_{t+1}}, \ldots, 0\}$, where $v_{t+1}$ is the next viewpoint. The

---

[1] https://www.open3d.org/docs/0.6.0/python_api/open3d.geometry.voxel_down_sample.html

* Corresponding author.
E-mail address: jianchao.ci@wur.nl.

weakly-supervised loss $L^w$ is then computed as follows:

$$L^w = \sum_{i=1}^{M} a_i (g_i - \hat{g}_i)^2 \quad (4)$$

where

$$a_i = \begin{cases} 1, & \text{if } i = v_{t+1} \\ 0, & \text{otherwise.} \end{cases} \quad (5)$$

This method allows the loss to be calculated only based on the ground truth of the next viewpoint $v_{t+1}$, significantly enhancing the efficiency of training data collection. However, it may lead to reduced sample efficiency, as only information associated with the next viewpoint is utilized for loss calculation and network optimization. To improve sample efficiency, experience replay is utilized (details refer to step 5 of section 2.1.1).

## 2.2. Experimental setup

Three experimental scenarios were designed to evaluate the performance of our SSL-NBV algorithm:

(1) Simulated Scenario 1 (experiment S1): This scenario evaluates the network's performance in reconstruction efficiency and quality, IG prediction speed, and training efficiency. We compared our SSL-NBV algorithm with other NBV and non-NBV methods. (Q1).
(2) Simulated Scenario 2 (experiment S2): Building on S1, this scenario modifies viewpoint settings and target plants to test the method's generalization and adaptability to novel view settings and target plants. In this experiment, We compared weakly-supervised and strongly-supervised learning (Q2), and analyzed performance improvements after online fine-tuning (Q3). The network trained in S1 was used as the starting point for fine-tuning in this experiment.
(3) Real-World scenario (experiment RW): This scenario involved testing the algorithm in real-world conditions using a robot equipped with an RGB-D camera and real plants (Q4). The neural network trained in S1 was continuously fine-tuned in this scenario.

To improve generalization, multiple plants were used during training. The entire training process was separated into many plant reconstruction cycles. Each cycle, a plant with random pose was positioned as the target, and the robot consecutively selects $n$ views from $C$ to reconstruct the plant, collecting training data and updating the network simultaneously. Once a reconstruction was complete after $n$ iterations, a new plant was positioned, the accumulated point cloud and the view-selection state were re-initialized, and a new cycle began. Through preliminary tests, $n = 10$ was set for simulation experiments, where we systematically increased the value, and 10 viewpoints typically allowed good plant reconstruction for our and the baseline methods. For the real-world experiment, $n = 15$ was set due to the higher complexity of real plants. The same value of $n$ was used during testing.

### 2.2.1. Simulation scenarios

The simulation was developed using Gazebo (Koenig and Howard, 2004), with data collection and exchange via the Robotic Operating System (ROS) (Quigley et al., 2009). The simulation ran on a ThinkPad P15 laptop with an Intel Xeon W-11855M CPU and an Nvidia GeForce RTX A500 GPU with 16 GB memory, operating on Ubuntu 20.04.

In experiment S1 and S2, as illustrated in Fig. 4, the candidate viewpoints set $C$ comprised 33 viewpoints ($M$=33), arranged in a cylindrical pattern around the origin of the global frame, providing observations of the plant from 11 angles ($a_1, ..., a_{11}$) and 3 heights ($h_1, h_2, h_3$). These viewpoints were all horizontal and oriented to face the Z axis of the origin. The camera moved freely between viewpoints, providing a 360° view of the plant. To create additional variation, each time a plant was created, a plant model was randomly selected from the plant model set and positioned randomly with coordinates $d_x$ and $d_y$ along the x and y axes, and a rotation $\theta$.

\* Corresponding author.
E-mail address: jianchao.ci@wur.nl.

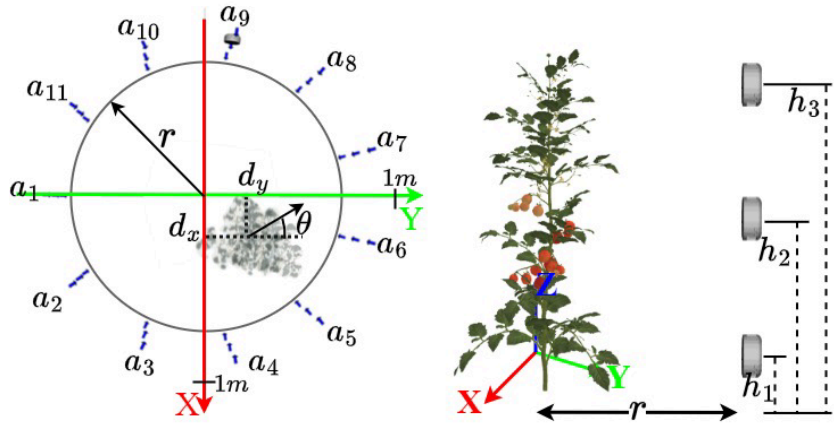

Fig. 4. Illustration of viewpoint sampling and plant creation. The left image depicts a top view and the right image depicts a side view. Viewpoints (blue arrows) were sampled in a cylindrical distribution to observe the plant from 11 angles ($a_1, ..., a_{11}$) and 3 heights ($h_1, h_2, h_3$). Each plant is created around the origin with random positions $d_x$ and $d_y$ in x and y axes, and a rotation $\theta$.

In experiment S1, the candidate viewpoints featured a radius $r$ of 0.6m, with different heights $h_1$=0.04m, $h_2$=0.25m, and $h_3$=0.46m. For the random translation, the range for $d_x$ and $d_y$ selections were set at $d_x, d_y \in U(-0.1, 0.1)$ with a 0.02m interval, and for the rotation was $\theta \in U(0, 360)$ with a 20º interval. Ten 3D tomato plant models[2] (T1-T10 as shown in Fig. 5) were used as targets, exhibiting variations in architecture, size, height, the number of tomato trusses, and leaf nodes. Eight plants were used for training and two for testing. Due to the limited number of plants, K-fold cross-validation was employed to evaluate the method. Those ten tomato plants were divided into two classes, A (T1-T5) with simpler structures and B (T6-T10) with more complex structures. The ten plants were then randomly split into 5 sets (K=5), each containing one plant from each class: set 1 (T3, T8), set 2 (T4, T6), set 3 (T5, T9), set 4 (T1, T7), and set 5 (T2, T10). This setup resulted in five rounds of validation, where each round used one set for testing and the others for training. Each validation round involved 50 repetitions of plant reconstruction for the testing set, totaling 250 repetitions (5 sets × 50 repetitions). Each repetition randomly selected a plant from the test set, placed it in a random location and orientation, and had the trained network reconstruct the plant starting from a random viewpoint. We chose 50 repetitions to ensure sufficient variability in the reconstruction process to reflect the method's overall performance.

---

[2] https://www.cgtrader.com/3d-models/plant/other/xfrogplants-tomato

[3] https://www.cgtrader.com/3d-models/plant/other/chili-pepper

\* Corresponding author.
E-mail address: jianchao.ci@wur.nl.

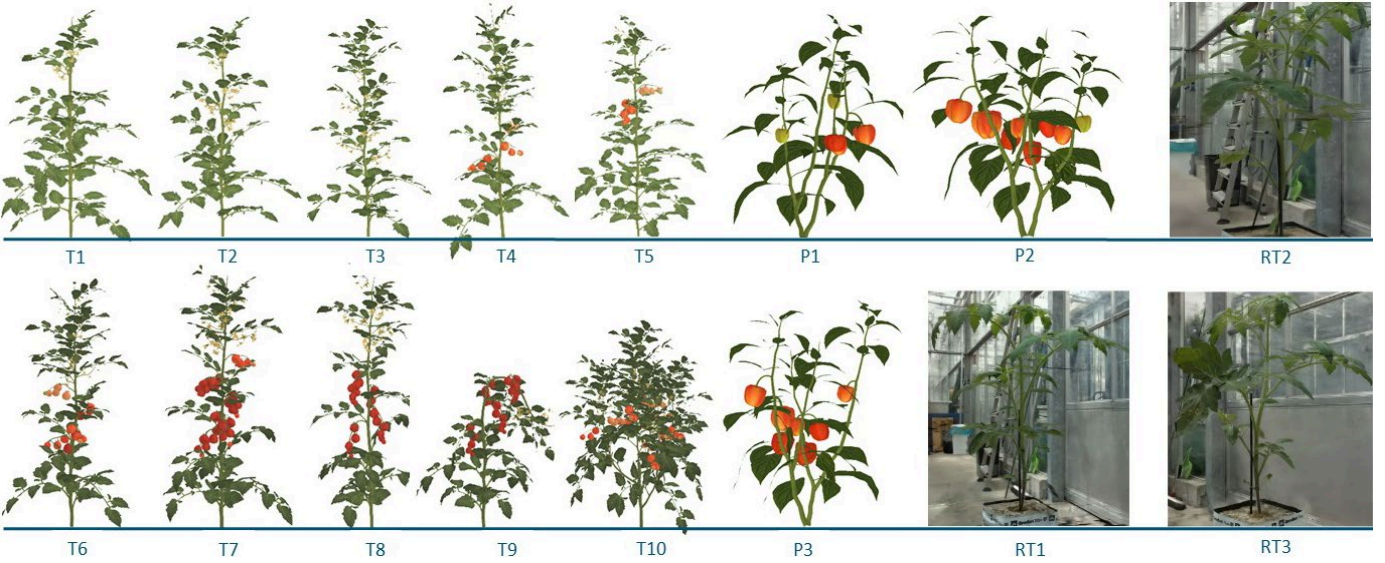

Fig. 5. Plants used in experiments. T1-T10 are simulated tomato plants used in experiment S1. P1-P3 are simulated pepper plants used in experiment S2. RT1-RT3 are real tomato plants used in experiment RW.

In experiment S2, the viewpoint settings and target plants were modified to create a novel environment, testing the online learning and fine-tuning capabilities of the proposed method in a completely novel scenario. The radius ($r$) was set at 0.5m, and a larger range of random plant positions $d_x, d_y \in U(-0.3, 0.3)$ was used to introduce more variation in the relative pose between the camera and the plant compared to experiment S1. Three pepper plants[3] (P1-P3 as shown in Fig. 5), significantly different in morphology from tomato plants were used. A K-fold cross-validation method with K=3 was employed during the evaluation. In each round of validation, one plant was used for testing while the remaining two were used for fine-tuning. Each validation round involved 50 repetitions, totaling 150 repetitions (3 plants × 50 repetitions).

### 2.2.2. Real-world scenario

The real-world setup involved an ABB IRB 1200 robot equipped with an Intel Realsense L515 RGB-D camera attached to its end effector (refer to Fig. 6a), facilitating flexible data collection from various viewpoints. The system was controlled by ROS for robot control and data communication between devices, using the same ThinkPad P15 laptop as simulation.

Due to the motion limitations of the robot, we sampled $M$=33 viewpoints in a semi-cylindrical distribution, spanning from -60° to 60° relative to the x-axis of the robot frame (see Fig. 6b). The radius of the cylindrical sector was 0.45m, with viewpoints at heights of $h_1$=0.75m, $h_2$=1.0m, and $h_3$=1.25m relative to the origin of robot frame. We selected three 40-day-old tomato plants (RT1-RT3, as shown in Fig. 5) as targets. These plants were approximately 55cm in height and had 6-7 composite leaves. The data from the real plants was pre-collected, allowing us to repeat experiments and compare the results with baseline methods.

For each plant, the position was fixed at 0.9m, 0m, and 0.65m along the x, y, and z axes of the robot frame, respectively. Four rotations of 0° or 90° or 180° or 270° were applied, and the point clouds for all 33 views for each rotation were collected. A k-fold cross-validation method was used to test the approach, with each round of validation using two plants as the training set and one plant as the testing set, resulting in a total of three rounds (k=3) of validation. Each round involved 50 repetitions, leading to a total of 150 repetitions (3 plants × 50 repetitions) of plant reconstruction. In each repetition, the test plant, set to one of the four rotations, was selected as the target, and the method reconstructed the plant from a randomly selected initial view.

* Corresponding author.
E-mail address: jianchao.ci@wur.nl.

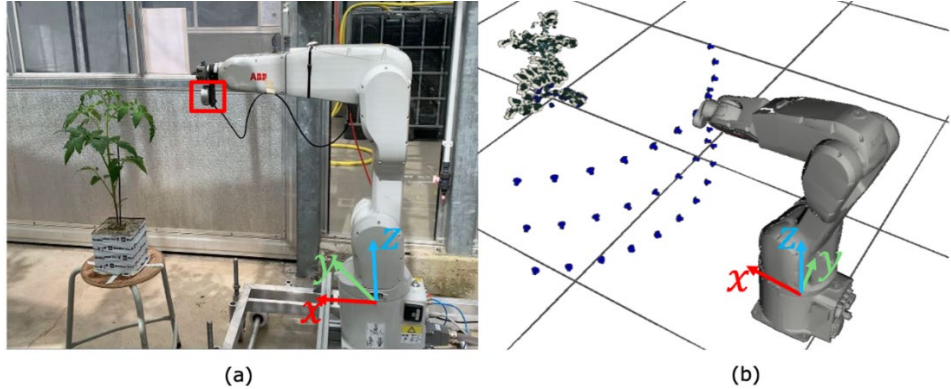

Fig. 6. Illustration of real-world setup. The real-world environment includes a real robot equipped with an RGB-D camera for flexible plant reconstruction.

### 2.2.3. Implementation details of Self-Supervised Learning in experiments

In all three experiment scenarios, during network training, a batch size $N=32$ for experience replay was set, and the buffer size was set to $l=1000$. Before feeding the accumulated point cloud $P_t^a$ to the network, a downsampling procedure was conducted to resize the point cloud to 512 points, independent of the size of the plants, with each point being randomly chosen from $P_t^a$. This choice was based on a preliminary test which indicated that using more points did not improve the network's accuracy in IG prediction and reduced prediction speed and training efficiency and increasing computation and memory demands due to the increased complexity of the network architecture.

In experiment S1, the exploration parameters were set to $\epsilon_{ini}$=1.0 and $\epsilon_{min}$=0.2, with the maximum number of training iterations set to $T$=50,000. In experiment S2, $\epsilon_{ini}$=0.2 was reduced to 0.2 since the network was already pre-trained in experiment S1, and $T$ was set to 12,400 to maintain the same number of iterations per plant in the training set (8 for tomato and 2 for pepper). In Experiment RW, $\epsilon_{ini}$ was set back to 1.0 to allow more exploration due to the substantial differences between simulation and real-world conditions. The maximum iterations were set to $T$=25,000, which was higher than in experiment S2, considering the increased complexity of real-world plants.

### 2.2.4. Implementation of baseline planners for experiments

We compared the performance of our SSL-NBV with other methods:

(1) Traditional Voxel-based NBV (Burusa et al., 2022): This method converts the point cloud into an occupancy grid and uses ray casting from each candidate viewpoint to calculate the IG for the next-best view. To enhance the speed of view planning, the occupancy grid updates and ray casting processes are executed using GPU acceleration. The parameters used for ray casting, IG calculation and occupancy calculation follow the original paper.
(2) PC-NBV (Zeng et al., 2020): This approach has a training setup similar to our SSL-NBV in terms of batch size, training iterations, and epochs. However, PC-NBV relies on strongly-supervised data, requiring the collection of ground-truth IG by traversing all candidate viewpoints, a process that is highly time-consuming and dependent on complete plant models, which are not available in real-world scenarios.

* Corresponding author.
E-mail address: jianchao.ci@wur.nl.

(3) Random planner: In this method, each next viewpoint is randomly selected from the candidate set $C$.

(4) Pre-defined planner: This planner selects 11 viewpoints from the candidate set $C$ and distributed in a zigzag pattern across 3 heights and 11 angles, alternating heights at each angle. The set of views is formulated as $\{a_1h_1, a_2h_2, a_3h_3, a_4h_2, a_5h_1, a_6h_2, a_7h_3, a_8h_2, a_9h_1, a_{10}h_2, a_{11}h_3\}$ (refer to Fig. 4). During each cycle of plant reconstruction, 10 viewpoints were randomly selected without repetition from this set and visited in random order to remove the impact of viewpoint order on plant reconstruction efficiency.

### 2.3. Evaluation metrics

Two main metrics were employed to evaluate the proposed method: (a) Reconstruction quality and efficiency; (2) Required number of ground-truth annotations for network training.

#### 2.3.1. Reconstruction quality and efficiency

Reconstruction quality was assessed using the reconstruction ratio ($R$) for the ground-truth point cloud $P^o$. In the simulation environment, the ground-truth point clouds of plant models were generated by uniformly sampling points on their mesh surfaces. These points were then down sampled using the VoxelGrid filter, resulting in a spatial resolution of 0.003m. In the real-world scenario, where 3D models were not available, we approximated the ground-truth model by merging the point clouds collected from all viewpoints. Subsequently, the same noise filtering process described in section 2.1.3 was applied, and the ground-truth point cloud was downsampled to a resolution of 0.01m. While the ground-truth generated using this method is approximate, it remains effective, considering that all planners are implemented with the same criteria.

The reconstruction ratio at any current reconstruction round was computed as the proportion of points within $P^o$ that were reconstructed in the current accumulated point cloud, expressed as $R = |P^o \cap P^a|/|P^o|$, where the intersection $P^a \cap P^o$ was calculated using Eq. (2). In experiment S1 and S2, the threshold $\delta$ was set at 0.003m to determine if a point in $P^o$ is reconstructed, while in RW, the threshold was set to 0.01m due to the additional sensor noise. Reconstruction efficiency was assessed by counting the number of views required to achieve specific thresholds $\tau$=0.8 and 0.9 of $R$ in a plant reconstruction cycle.

The IG prediction speed can significantly impact reconstruction efficiency. A faster IG prediction method enables faster next-best-view selection, leading to faster plant reconstruction. IG prediction speed was measured as the time taken to predict the IG for all 33 candidate views, with and without GPU acceleration.

The performance of our SSL-NBV algorithm was evaluated in comparison to baseline planners using K-fold cross-validation. During each round of validation, all planners were applied to the same set of test plants and performed plant reconstruction with the same number of repetitions (details are provided in sections 2.2.1 and 2.2.2). The results for each method were obtained by averaging the outcomes from these K-fold cross-validation repetitions.

#### 2.3.2. Number of ground-truth annotations

The measurement of each ground-truth IG required the robot to move its end-effector to a candidate view, collect a partial point cloud, and calculate IG based on the accumulated point cloud. As a result, the number of ground-truth annotations necessary for network training significantly impacted the adaptability of online learning methods. A method that requires fewer annotations can reduce robotic motion, accelerate data collection, and facilitate more frequent network updates.

The required number of annotations for SSL-NBV, denoted as $A^{ssl} \in \mathbb{N}$, was calculated equaling the total number of robotic motions, as it used online weakly-supervised learning, each robotic motion allowed the collection of a training sample and its corresponding ground-truth IG. To give a direct insight about the results, the value of $A^{ssl}$ was compared with PC-NBV, which was trained using offline strongly-supervised learning. The required number of annotations for PC-NBV was calculated as $A^{off} = F \times M$, where $F$ represents the total number of offline training samples and $M$ denotes the number of candidate views.

* Corresponding author.
E-mail address: jianchao.ci@wur.nl.

Since the training data for PC-NBV was pre-collected, $A^{off}$ remained fixed throughout the entire training process. We used $A^{off}$ as the maximum allowable number of annotations for SSL-NBV during online training. As training progressed, SSL-NBV's performance in plant reconstruction improved, and $A^{ssl}$ increased until reaching $A^{off}$. Practically, SSL-NBV's training could be terminated when its plant reconstruction performance converged to a maximum level, as the network's weights stabilize and optimize. However, determining the exact convergence point can be subjective. To address this, we proposed a statistical approach where training was considered complete when there was no significant difference (p-value > 0.05) between the reconstruction ratio of SSL-NBV and PC-NBV. This point was then used to calculate the $A^{ssl}$ for SSL-NBV.

In experiment S1, 8,000 samples were collected for PC-NBV training, resulting in $A^{off}$=264,000(8000 samples × 33 views). In experiment RW, 1,920 samples were collected for fine-tuning PC-NBV, resulting in $A^{off}$=63,360(1920 samples × 33 views).

## 3. Results

### 3.1. Simulated scenario 1

The results of experiment S1 addressed Question 1 by comparing the plant reconstruction efficiency and quality of SSL-NBV with baseline methods. Fig. 7 shows the average reconstruction ratio of SSL-NBV compared to other methods over cross-validation. All NBV methods outperformed non-NBV methods, confirming that NBV approaches are more effective for plant reconstruction. Our method achieved a final reconstruction ratio of approximately 0.95, with $\tau$=0.8 and 0.9 reached after 5 and 6 viewpoints, respectively. This was faster than the non-NBV methods, where the Pre-defined planner required 5 and 7 views, and the Random planner needed 7 and 10 views, indicating that SSL-NBV can more efficiently reconstruct plants with fewer views. Compared to PC-NBV, which was trained using strongly-supervised learning, SSL-NBV showed a slightly lower reconstruction efficiency, with a 0.02 reduction in the final reconstruction ratio. However, SSL-NBV uses weakly-supervised learning, requiring only sparse IG labels, can significantly reduce the need for ground-truth IG annotations during training compared to PC-NBV (further analysis is provided subsequently). None of the planners achieved 100% plant reconstruction because the viewpoints were restricted to horizontal views, missing parts of the plant only visible from other angles, such as looking up or down.

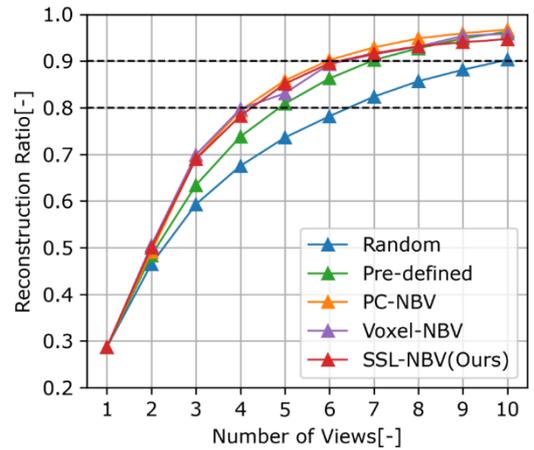

Fig. 7. Average reconstruction ratio of SSL-NBV in experiment S1 across cross-validation (250 plant reconstruction cycles) compared to baseline methods.

Fig. 8 shows the plant reconstruction ratio of SSL-NBV as a function of the number of annotations used during online learning. This is a representative curve with T1-T7 as test plants, and a similar pattern was observed in other validation rounds. The figure compares weakly-supervised (SSL-NBV) and strongly-supervised (PC-NBV) learning in terms of sample efficiency, partially addressing Question 2. To compare to PC-NBV, which used 264,000 annotations during training. In this comparison, the training of SSL-NBV was also extended to 264,000 training iterations but the plot shows that already with fewer annotations, SSL-NBV reaches similar performance. The results indicate that SSL-NBV achieved a similar plant reconstruction ratio to PC-NBV (p-value>0.05) with only 9% of the annotations, indicating a significantly higher sample efficiency. This increased sample efficiency can be attributed to the different training approaches. PC-NBV used offline strongly-supervised learning, updating the network to improve

* Corresponding author.
E-mail address: jianchao.ci@wur.nl.

IG prediction accuracy for all views simultaneously. While this approach could enhance generalization and reduce fluctuations in training, it could also introduce redundant or less informative data, reducing sample efficiency. Conversely, SSL-NBV employed weakly-supervised learning with a balance between exploration and exploitation, which likely results in the selection of more informative views.

While SSL-NBV exhibited a slightly lower average reconstruction ratio than PC-NBV when $A^{ssl} = A^{off}$, the difference was not statistically significant, suggesting that, given the same number of annotations, SSL-NBV can achieve similar plant reconstruction performance with PC-NBV. And consider the significant advantages of SSL-NBV in sample efficiency and online learning adaptability, it is well-suited for automatic plant reconstruction tasks.

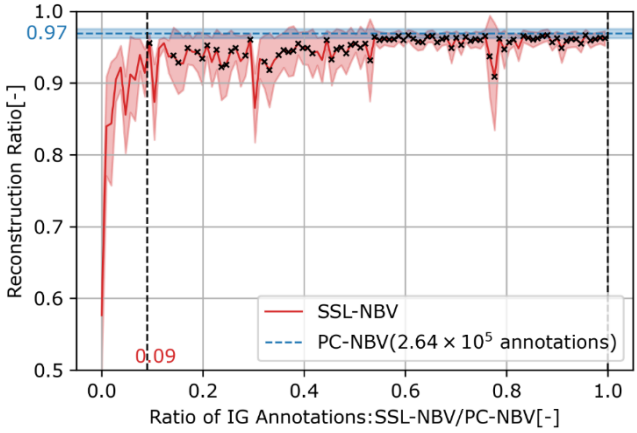

Fig. 8. The plant-reconstruction ratio as a function of the number of training annotations used during online learning, using T1-T7 as test plants. The x-axis shows the ratio of used IG annotations used by SSL-NBV compare to the $2.64 \times 10^5$ annotations used by PC-NBV. The red line shows the trend for the SSL-NBV method, while the blue line for the PC-NBV method is plotted as a reference. The shaded regions represent the 95% confidence interval over 50 plant reconstruction cycles. Crosses (x) indicate reconstruction ratios not significantly different (p-value>0.05) between SSL-NBV and PC-NBV. The vertical black dashed line indicates the first ratio of IG annotation where the two methods do not show a significant difference in the reconstruction ratio.

Furthermore, we compared the IG prediction speed between our SSL-NBV, PC-NBV, and Voxel-NBV, as displayed in Table 1. This speed was measured by the total time required by each method to predict the IG for all candidate views. Since SSL-NBV and PC-NBV used the same network architecture, they achieved the same prediction speed. When the GPU was utilized, our SSL-NBV required only 0.0039 seconds for IG prediction, representing a substantial 818-times improvement over Voxel-NBV (3.19s). This highlights the efficiency gain of neural-network-based NBV methods compared to voxel-based approaches that use ray-casting for IG prediction. When only the CPU was available, our method required 0.03 seconds for IG prediction, achieving a 733-times improvement over Voxel-NBV (22s). Importantly, these time measurements considered only IG prediction. For Voxel-NBV, which requires additional time for converting point cloud data to voxels, the difference in speed between the two methods would be further increased.

Table 1. Comparison of IG prediction speed between the learning-based NBV methods and the voxel-based method with ray-casting.

| GPU/CPU | Planner | Step Time (s) (Average±Std) |
|---|---|---|
| GPU | SSL-NBV/PC-NBV | 0.0039±0.00047 |
| | Voxel-NBV | 3.19±0.0038 |
| CPU | SSL-NBV/PC-NBV | 0.03±0.0047 |
| | Voxel-NBV | 22±1.8 |

### 3.2. Simulated scenario 2

The results of experiment S2 addressed Question 2 by comparing the generalization capability of weakly-supervised learning and strongly-supervised learning, and Question 3 by testing the adaptability of the proposed method in a novel environment. Fig. 9 shows the average reconstruction ratio of SSL-NBV compared to other NBV methods over cross-validation. In this experiment, the SSL-NBV network trained in experiment S1 with $T$=50,000 iterations on tomato plants was continuously fine-tuned for pepper plants. The fine-tuned SSL-NBV demonstrated improved reconstruction efficiency and quality compared to the non-fine-tuned SSL-NBV network, reaching τ=0.8 and 0.9 with 1 and 2 fewer views, respectively, showing its adaptability to automatically optimize for new scenarios, addressing Question 3. The fine-tuned network reached τ=0.8 with 1 fewer viewpoint than PC-NBV. However, both methods achieved comparable reconstruction at the end, indicating that PC-NBV has a high level of generalization in this novel environment and can produce good plant reconstruction with sufficient viewpoints.

* Corresponding author.
E-mail address: jianchao.ci@wur.nl.

The non-fine-tuned SSL-NBV showed lower reconstruction efficiency than PC-NBV, indicating that weakly-supervised learning offers lower generalization compared to strongly-supervised learning. However, SSL-NBV was trained with significantly fewer annotations (50,000 for SSL-NBV compared to 264,000 for PC-NBV) than PC-NBV, still achieved a similar final reconstruction ratio (addressing Question 2). After automatic online fine-tuning, SSL-NBV outperformed PC-NBV. Although the improvement was small, we attribute this to the fact that experiment S2 still shared some similarities with experiment S1, enabling PC-NBV to generalize to this environment. The Voxel-NBV yielded the poorest results, possibly because the grid space was enlarged to allow more variation in plant position in this scenario (refer to section 2.2.1). This introduced more empty voxels, impacting the performance of Voxel-NBV, as this method aimed to reconstruct the entire grid space rather than focusing solely on the plant.

### 3.3. Real world

The results of experiment RW addressed Question 4 by evaluating the SSL-NBV method in a real-world scenario. Fig. 10 illustrates the average reconstruction ratio of SSL-NBV compared to other methods over cross-validation. SSL-NBV showed effectiveness in this real-world scenario, outperforming non-NBV planners in both reconstruction quality and efficiency. Specifically, SSL-NBV achieved a τ =0.8 within 7 viewpoints, which is 2 and 3 views faster than the Random and Pre-defined planners, respectively. SSL-NBV reached a τ =0.9 after 12 views, a threshold that the non-NBV methods failed to reach. SSL-NBV reconstructed approximately 92.5% of the plant, representing a 3.5% and 8.5% improvement over the Random and Pre-defined planners, respectively.

Notably, in this experiment, we approximated the ground-truth plant model by merging point clouds from all viewpoints, which made the ground-truth IG calculation and offline fine-tuning of PC-NBV possible. However, in real-world applications, ground-truth plant models are not available for PC-NBV fine-tuning. In contrast, our SSL-NBV does not require ground-truth plant models and automates the entire training/fine-tuning process to collect its own training data and to continuously optimize the network to adapt to novel plants and environments in a self-supervised manner.

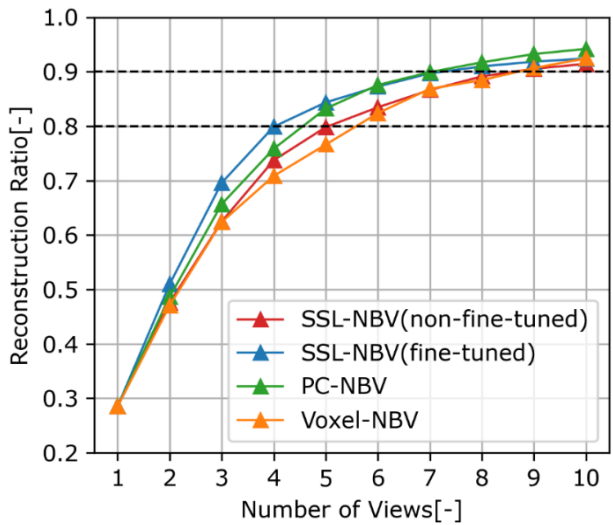

Fig. 9. Average reconstruction ratio of SSL-NBV in experiment S2 across cross-validation (150 plant reconstruction cycles) compared to baseline methods.

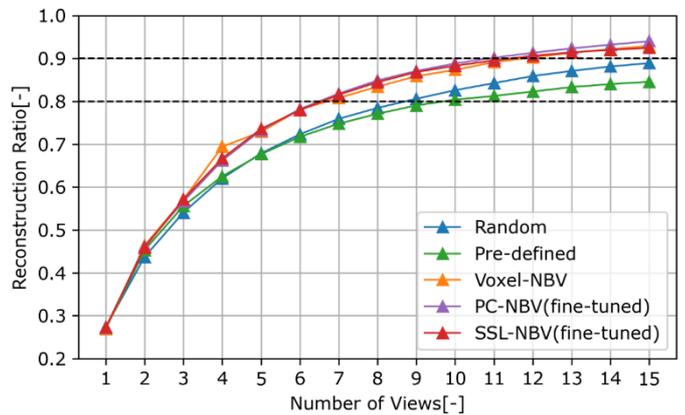

Fig. 10. Average reconstruction ratio of SSL-NBV in experiment RW evaluated using cross-validation (150 plant reconstruction cycles) compared to the baseline methods.

Fig. 11 shows the trend of SSL-NBV's reconstruction ratio as an increased number of annotations were used during online learning, with RT3 as the test plant. During fine-tuning, 1,920 samples were recollected to retrain PC-NBV, resulting in $A^{off}$ = 63,360 IG annotations. To align with PC-NBV (with 63,360 annotations), the training of SSL-NBV was extended to 63,360 training iteration. The figure shows SSL-NBV reached a similar reconstruction ratio (p-value>0.05) with PC-NBV using only 21% of the annotations, demonstrating a 79% reduction in the need for ground-truth annotations.

* Corresponding author.
E-mail address: jianchao.ci@wur.nl.

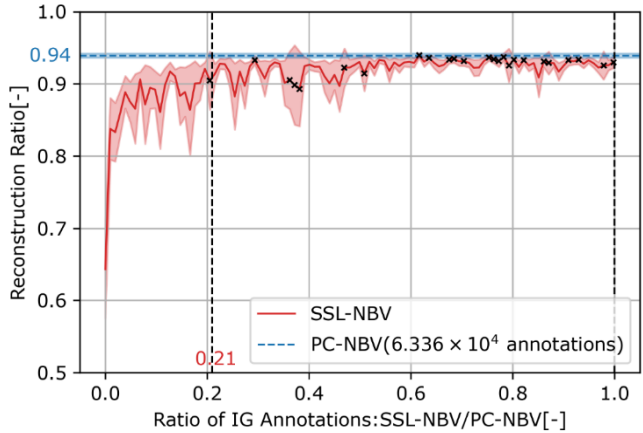

Fig. 11. The plant-reconstruction ratio as a function of the number of training annotations used during online learning, using RT3 as the test plant. The shaded regions represent the 95% confidence interval over 50 plant reconstruction cycles. The x-axis shows the ratio of IG annotations used by SSL-NBV over PC-NBV. Crosses (x) indicate reconstruction ratios not significantly different (p-value>0.05) between SSL-NBV and PC-NBV.

## 4. Discussion

### 4.1. Comparison with relevant studies

The results demonstrated that learning-based NBV methods achieved plant reconstruction performance comparable to the classic Voxel-based NBV method, while significantly improving IG prediction speed. This aligns with Han et al. (2022), who developed a Double Branch NBV Network (DB-NBV) for 3D object reconstruction and evaluated it on synthetic datasets. Our findings also support Zapotezny-Anderson and Lehnert (2019), who showed that using a deep neural network for direct IG prediction improved viewpoint planning efficiency.

Compared to PC-NBV, which relied on strongly-supervised learning with fully labelled IG data, our method required only sparse IG labels and achieved similar plant reconstruction efficiency while significantly reducing the number of ground-truth IG annotations. Our findings are consistent with Stutz and Geiger (2020), who demonstrated that using 3-10% of annotations can yield comparable results to fully annotated training data in 3D shape completion tasks. Similarly, Cheng et al. (2023) showed that using weakly-supervised labels can achieve comparable performance to fully labelled data in image segmentation tasks. To our knowledge, the application of weakly-supervised learning to NBV systems has not been explored, with existing approaches like those of Han et al. (2022), Zeng et al. (2020) and Mendoza et al. (2020) relying on extensive IG labels.

Notably, our method has the capability to automatically fine-tune for novel plants and environments without human intervention. When integrated with robotics, it enabled life-long and continuous online learning, which is crucial for robotic agricultural operations where plants and environments are constantly changing. This capability is absent in existing learning-based NBV methods, which typically rely on pre-collected offline data for training.

We conducted a quantitative analysis of the method's performance in real-world plant reconstruction, an important evaluation lacking in similar studies. While Han et al. (2022) applied their method to real-world plant reconstruction, their evaluation was qualitative and lacked quantitative comparison. Furthermore, their approach required complete object models for ground-truth IG calculation, limiting its ability to be fine-tuned for real-world plants where such models are unavailable. This issue was also present in the works of Zeng et al. (2020) and Mendoza et al. (2020). Instead, our method addressed this by improving the IG metric and eliminating the need for complete plant models in IG calculation, enabling continuous fine-tuning for real-world plants.

In summary, this study presented SSL-NBV, a method capable of automatically adapting to novel plants and environments, allowing for life-long and continuous improvement of the robot during task execution without human intervention. Through comprehensive evaluation in both simulation and real-world experiments, we demonstrated its efficiency in IG prediction and its ability to significantly reduce the need for IG annotations. This work offers valuable insights for future applications of NBV methods in agriculture, particularly in reducing the reliance on extensive annotations and enhancing scalability in dynamic environments.

* Corresponding author.
E-mail address: jianchao.ci@wur.nl.

### 4.2. Limitations and future improvements

#### 4.2.1. Flexible viewpoint sampling for different-sized plants

Our method used a global viewpoint sampling approach, where a set of candidate viewpoints were pre-sampled, and the NBV planner iteratively selected from this set to reconstruct the plant. This approach is efficient, as it enables a global search for the next best viewpoint. However, it restricted the selection to fixed viewpoints, potentially missing optimal views and reducing reconstruction quality. Additionally, the use of a fixed set of candidate views can limit the method's scalability for different plant sizes. Zeng et al. (2022) and Lehnert et al. (2018) addressed this challenge with relative viewpoint planning, where the next camera pose was determined based on incremental movements from the current view. These methods are not restricted to fixed viewpoints and are adaptable to various plant sizes. However, local view planning methods often face issues with local maxima. In future work, we aim to integrate SSL-NBV with local NBV methods, facilitating global planning for large-scale search (e.g., entire plant or row of plants) while enabling detailed reconstruction of plant parts not visible from fixed viewpoints.

#### 4.2.2. Trajectory optimization of camera view

Our approach performed NBV planning on a viewpoint-wise manner, selecting the next best view solely based on the maximization of IG. However, this can lead to low view trajectory efficiency, requiring extensive camera movement to achieve a similar level of plant reconstruction. Ideally, NBV planning should consider view trajectory optimization, i.e., minimizing camera movement while maintaining high reconstruction efficiency. To address this, we propose integrating a trajectory optimization mechanism into the existing NBV planning process using a DRL algorithm. DRL is widely used in path planning as it considers long-term impacts, learning a policy about task execution that maximizes cumulative rewards. In the NBV problem, incorporating both reconstruction efficiency and camera movement into the reward function could enable efficient plant reconstruction while reducing camera movement. Another potential solution is to add constraints to the current view selection process, allowing only nearby views to be selected after weighing camera movement against the predicted IG.

### 4.3. Impact of experimental conditions

#### 4.3.1. Size and distribution of set of candidate viewpoints

The size of the candidate set can affect the network's training efficiency and the quality of plant reconstruction. Increasing the candidate set size can improve reconstruction by capturing previously unseen plant parts but also reduces training efficiency and increases the need for training data. In this study, we used a set of 33 viewpoints, which balanced plant reconstruction and training efficiency. However, for larger plants or row of plants reconstruction tasks, expanding the candidate set may be necessary to fully capture all plant parts. A larger candidate set can amplify the advantages of NBV methods over non-NBV methods in plant reconstruction, as it becomes more difficult for non-NBV methods to select high-IG viewpoints, while NBV approaches, given sufficient training (for DL-NBV) or time (for Voxel-NBV), can effectively identify the next-best viewpoint, improving plant reconstruction results.

In this study, we employed cylindrical and semi-cylindrical distributions in simulation and real-world environments, respectively, similar to Golbach et al. (2016). This configuration facilitated plant reconstruction because the plants were centrally located so that each viewpoint had a high probability of capturing a plant part. However, in terms of evaluation, this setup may narrow the gap between NBV and non-NBV methods by making it easier for non-NBV methods to select viewpoints with high IG.

* Corresponding author.
E-mail address: jianchao.ci@wur.nl.

### 4.3.2. Limited real-world plants

Three real plants were used in this study to evaluate the method's performance in real-world scenarios. While testing on only three plants may not fully evaluate the method's overall performance due to the limited coverage of plant variations, we believe it still demonstrated the feasibility of the method for real plant reconstruction. Firstly, data for each plant were collected from four different poses, adding variations to the training and testing sets. Secondly, comprehensive evaluations were conducted in simulations with a wide range of plant variations, which helped validate and support the real-world results. Thirdly, all results were obtained through k-fold cross-validation, which enhanced the robustness of the evaluation. We anticipate that testing on more real plants, the methods would yield similar patterns to those observed in our study. However, to thoroughly evaluate the method's performance in the real world, we recommend conducting further experiments that include a broader range of plants and plant variations.

Additionally, although we performed a three-step noise removal process on the point cloud, there was still significant noise, especially at the edges of the ground-truth point cloud. This noise reduced the reconstruction ratio, as noise present in the ground-truth point cloud may be absent in the reconstruction. To minimize the effect of noise, we set the threshold $\delta$ to 0.01m (refer to Eq. (2) for detailed explanation). Although this relatively large threshold may overestimate reconstruction performance by easily counting a point in the ground truth as reconstructed, it ensured fairness from a comparative perspective and did not affect the comparison between methods, as the same threshold was applied to all methods. In addition, the impact of camera noise can be decreased in the future as technology advances and RGB-D cameras become more accurate and cheaper.

### 4.3.3. SSL-NBV training iterations

During SSL-NBV online training, we set $T$ =50,000, 12,400, and 25,000 iterations for experiments S1, S2, and RW, respectively. These settings allowed the method to achieve comparable performance in plant reconstruction to PC-NBV while significantly reducing training time. However, as observed in Fig. 8 and Fig. 11, with additional training iterations, the plant reconstruction performance of SSL-NBV could be further improved, bringing it closer to that of PC-NBV. Thus, we believe that the results in Fig. 7, Fig. 9, and Fig. 10 of SSL-NBV could also show improved performance if extended training iterations were applied, potentially reaching similar results with PC-NBV.

## 5. Conclusion

This paper proposed an SSL-NBV algorithm to enhance learning-based NBV methods through self-supervised learning, enabling efficient and automatic training of neural networks to predict the IG of next viewpoints, which allows to optimize 3D plant reconstruction. Using a robot to actively change viewpoints, the method facilitated life-long and continuous online learning without requiring human annotations. To ensure efficient online learning, an improved IG metric, experience replay, and weakly-supervised learning techniques were incorporated. A comprehensive evaluation with k-fold cross-validation was conducted to address the research questions.

The results of experiment S1 showed the reconstruction quality and efficiency for different multi-view reconstruction methods (answering Question 1). The proposed SSL-NBV method outperformed non-NBV methods, achieving the level of 90% plant reconstruction with fewer views, respectively 1 view and 4 views for the Pre-defined planner and the Random planner. The SSL-NBV method achieved similar reconstruction quality and efficiency compared to other NBV methods. Compared to the classic Voxel-based NBV, the proposed method achieved an 818 times higher IG prediction speed. Compared to PC-NBV, which used offline strongly-supervised learning, the proposed method achieved comparable plant reconstruction performance with only 9% of the IG annotations

\* Corresponding author.
E-mail address: jianchao.ci@wur.nl.

(answering Question 2). Moreover, this was achieved purely in a self-supervised way, without the need of human annotation or a ground-truth plant model. Question 3 was addressed in experiment S2. After online self-supervised learning, the method achieved 80% and 90% plant reconstruction 1 and 2 views faster than the non-fine-tuned network, respectively, demonstrating the method's adaptability to new environment though online self-supervised learning (answer Question 3). Also, a clear improvement in plant reconstruction during online learning (as shown in Fig. 8 and Fig. 11) further confirmed the method's adaptability. Finally, the results of experiment RW addressed showed that the proposed method could also successfully be applied in the real world (Question 4), achieved 92.5% plant reconstruction after online fine-tuning on real plants. It outperformed the Random and Pre-defined planners by around 3.5% and 8.5%, respectively, while reducing the need for IG annotations by 79% compared to PC-NBV.

In conclusion, the proposed SSL-NBV method was capable of efficient 3D plant reconstruction in simulated and real-world environments and could adapt to novel environments through online self-supervised learning without the need of any human intervention. Although this work was demonstrated using single plants, we believe that the proposed method can be adapted in future research for robotic plant reconstruction in more complex agricultural environments. Adaptations should overcome the current limitation of having a small number of fixed candidate views, and instead use more flexible view sampling and planning of view trajectories.

## CRediT authorship contribution statement

**Jianchao Ci:** Conceptualization, Methodology, Data curation, Software, Investigation, Writing-Original draft preparation, Formal analysis. **Eldert J. van Henten:** Conceptualization, Writing-Reviewing and Editing, Supervision, Funding acquisition; **Xin Wang:** Conceptualization, Writing-Reviewing and Editing, Supervision; **Akshay K. Burusa:** Conceptualization, Writing-Reviewing and Editing, Data curation, Software; **Gert Kootstra:** Conceptualization, Writing-Reviewing and Editing, Supervision, Funding acquisition.

* Corresponding author.
E-mail address: jianchao.ci@wur.nl.

## Declaration of competing interest

The authors declare that they have no known competing financial interests or personal relationships that could have appeared to influence the work reported in this paper.

## Data availability

Data will be made available on request.

## Acknowledgements

This work was supported by the China Scholarship Council (No.202107720034); and the Netherlands Organisation for Scientific Research (NWO) project "FlexCRAFT: Cognitive Robots for Flexible Agro-Food Technology", grant P17-01. We thank the members of the Agricultural Biosystems Engineering (ABE) group at Wageningen University for their insightful discussions and valuable feedback.

* Corresponding author.
E-mail address: jianchao.ci@wur.nl.

* Corresponding author.
E-mail address: jianchao.ci@wur.nl.